\documentclass[11pt, a4paper]{article}

\usepackage{amsmath}
\usepackage{amsfonts}
\usepackage{amssymb}
\usepackage{graphicx,color}
\usepackage{url,cite}
\usepackage{algorithm}

\newcommand{\beqar}{\begin{eqnarray}}
\newcommand{\eeqar}{\end{eqnarray}}
\newcommand{\beqarno}{\begin{eqnarray*}}
\newcommand{\eeqarno}{\end{eqnarray*}}
\newcommand{\ba}[1]{\begin{array}{#1}}
\newcommand{\ea}{\end{array}}
\newcommand{\rr}{{\mathbb R}}
\newcommand{\NN}{{\mathcal N}}

\newcommand{\QQ}{{\mathcal Q}}
\newcommand{\smallmat}[1]{\left[\hspace*{-.2em} \begin{smallmatrix}#1 \end{smallmatrix} \hspace*{-.2em}\right]}
\newcommand{\st}{\mathop{\rm s.t.}\nolimits}
\newcommand{\prox}{\mathop{\rm prox}\nolimits}
\newcommand{\diag}{\mathop{\rm diag}\nolimits}
\newcommand{\round}{\mathop{\rm round}\nolimits}
\newcommand{\eqdef}{\triangleq}
\newcommand{\matrice}[2]{\left[\hspace*{-.1cm}\ba{#1} #2 \ea\hspace*{-.1cm}\right]}

\newcommand{\tx}{\theta_x}
\newcommand{\ty}{\theta_y}
\newcommand{\txo}{\theta_{x0}}

\usepackage{enumitem}
\renewcommand{\theenumi}{\arabic{enumi}}

\renewcommand{\theenumii}{\arabic{enumii}}

\begin{document}

\title{Training Recurrent Neural Networks by 
    Sequential Least Squares and the Alternating Direction Method of Multipliers}

	\author{Alberto Bemporad\thanks{The author is with the IMT School for Advanced Studies, Piazza San Francesco 19, Lucca, Italy. Email: \texttt{\scriptsize alberto.bemporad@imtlucca.it}.
This paper was partially supported by the Italian Ministry of University and Research under the PRIN'17 project ``Data-driven learning of constrained control systems'', contract no. 2017J89ARP.}}

\maketitle
\thispagestyle{empty}

\begin{abstract}
This paper proposes a novel algorithm for training recurrent neural network models of nonlinear dynamical systems from an input/output training dataset. Arbitrary convex and twice-differentiable loss functions and regularization terms are handled by sequential least squares and either a line-search (LS) or a trust-region method of Levenberg-Marquardt (LM) type for ensuring convergence.
In addition, to handle non-smooth regularization terms such as $\ell_1$, $\ell_0$, and group-Lasso regularizers, as well as to impose possibly non-convex constraints such as integer and mixed-integer constraints, we combine sequential least squares
with the alternating direction method of multipliers (ADMM). We call the resulting algorithm
NAILS (nonconvex ADMM iterations and least squares) in the case
line search (LS) is used, or NAILM if a trust-region method (LM) is employed instead.
The training method, which is also applicable to feedforward neural networks as a special case,
is tested in three nonlinear system identification problems. 
\end{abstract}


\section{Introduction}
In the last few years we have witnessed a tremendous attention
towards machine learning methods in several fields, including control.
The use of neural networks (NNs) for modeling dynamical systems, 
already popular in the nineties \cite{HSZG92,SVD95}, is flourishing again,
mainly due to the wide availability of excellent open-source packages
for automatic differentiation (AD). AD greatly simplifies setting up and solving
the training problem via numerical optimization, in which the unknowns are the weights
and bias terms of the NN. While the attention has been mainly on the use
of \emph{feedforward} NNs to model the output function of a dynamical
system in autoregressive form with exogenous inputs (NNARX), 
\emph{recurrent} neural networks (RNNs) often better capture
the dynamics of the system, due to the fact that they are intrinsically 
state-space models.

While minimizing the one-step-ahead output prediction error of the NNARX model leads
to a (nonconvex) unconstrained optimization problem with a separable objective function, training RNNs is more difficult. This is because the hidden states and parameters of the network are coupled by nonlinear equality
constraints. The training problem is usually solved by gradient descent methods 
in which backpropagation through time (or its approximate truncated version~\cite{WP90})
is used to evaluate derivatives. To enable the use of minibatch stochastic gradient descent (SGD)
methods, in~\cite{BTS21} the authors propose to split the experiment in multiple sections
where the initial state of each section is written as a NN function of a finite set of past inputs and outputs.

The main issue of gradient-descent methods is their well-known slow-rate of convergence. 
To improve speed and quality of training, as well as to be able to train RNN models
in real-time as new input/output data become available, training methods based
on extended Kalman filtering (EKF)~\cite{PF94,WH11,LFB92} or
recursive least squares (RLS)~\cite{XKML94} were proposed based on
minimizing the mean-squared error (MSE) loss. These results were recently
extended in~\cite{Bem21} to handle rather arbitrary convex and twice-differentiable loss terms
when training RNNs whose state-update and output functions are
multi-layer NNs, showing a drastic improvement in terms of computation efficiency and quality of the trained model with respect to gradient-descent methods.
For the special case of shallow RNNs in which the states are known functions
of the output data and MSE loss, a combination of gradient descent and 
recursive least squares was proposed in~\cite{PDRO96} to optimize the sequence 
of outputs and weights, respectively.

In this paper we propose an offline training method based on
two main contributions. First, to handle arbitrary convex and twice-differentiable loss terms, we propose the use of sequential least squares,
constructed by linearizing the RNN dynamics successively and taking a quadratic approximation
of the loss to minimize. An advantage of this approach, compared to more classical backpropagation~\cite{Wer90} is that the required Jacobian matrices, whose evaluation is usually the most expensive part of the training algorithm, can be computed
in parallel, rather than sequentially, to propagate the gradients. 
To force the objective function to decrease monotonically, 
we consider two alternative ways: a line-search (LS) method and a trust-region method,
the latter in the classical Levenberg-Marquardt (LM) setting~\cite{Lev44,Mar63,TS12}.

A second contribution of this paper is to combine sequential least squares with nonconvex alternating direction method of multipliers (ADMM) iterations to handle non-smooth and possibly non-convex regularization terms. ADMM has been proposed in the past for training NNs, mainly feedforward NNs~\cite{TBXSPG16} but also recurrent ones~\cite{TKSQXLL20}. However, existing methods entirely rely on ADMM to solve the learning problem, such as to gain efficiency by parallelizing numerical operations. In this paper, ADMM is combined with the smooth nonlinear optimizer 
based on sequential LS mentioned above to handle non-smooth terms, 
therefore allowing handling sparsification (via $\ell_1$, $\ell_0$, or group-Lasso penalties) and to impose (mixed-)integer constraints, such as for quantizing the network coefficients. 
In particular, we show that we can use ADMM and group Lasso to select
the number of states to include in the nonlinear RNN dynamics, therefore
allowing a tradeoff between model complexity and quality of fit, a fundamental aspect
for the identification of control-oriented black-box models. See also~\cite{LDWWZKQXLQW19} for the use of nonconvex ADMM to handle combinatorial constraints in combination with RNN training.

We call the overall algorithm NAILS (nonconvex ADMM iterations and least squares)
when line search (LS) is used, or NAILM if a trust-region approach of Levenberg-Marquardt (LM) 
type is employed instead. We show that NAILS/NAILM is computationally more efficient,
provides better-quality solutions than classical gradient-descent methods, and is competitive with respect to EKF-based approaches. The algorithm is also applicable to the simpler case of training feedforward neural networks. 

The paper is organized as follows. After formulating the training problem in Section~\ref{sec:problem_formulation}, we present the sequential least-squares
approach to solve the smooth version of the problem in Section~\ref{sec:SLS}
and extend the method to handle non-smooth/non-convex
regularization terms by ADMM iterations in Section~\ref{sec:ADMM}.
The performance and versatility of the proposed approach is shown in three
nonlinear identification problems in Section~\ref{sec:experiments}.

\section{Problem formulation}
\label{sec:problem_formulation}
Given a set of input/output training data $(u_0,y_0)$, $\ldots$, $(u_{N-1},y_{N-1})$,
$u_k\in\rr^{n_u}$, $y_k\in\rr^{n_y}$, we want to identify a nonlinear state-space model in recurrent neural network (RNN) form
\begin{equation}
        x_{k+1}=f_x(x_k,u_k,\tx)
        ,\quad \hat y_{k}=f_y(x_k,u_k,\ty)
\label{eq:RNN}%
\end{equation}
where $k$ denotes the sample instant, $x_k\in\rr^{n_x}$ is the vector of hidden states,
$f_x:\rr^{n_x}\times\rr^{n_u}\to\rr^{n_x}$ and 
$f_y:\rr^{n_x}\times\rr^{n_u}\to\rr^{n_y}$ are feedforward neural networks parameterized by weight/bias terms $\tx\in\rr^{n_{\theta x}}$ and $\ty\in\rr^{n_{\theta y}}$, respectively, as follows~\cite{Bem21}:
\begin{subequations}
\begin{equation}
\left\{\ba{rcl}
    v_{1k}^x&=&A_{1}^x\smallmat{x_k\\u_k}+b_1^x\\
    v_{2k}^x&=&A_{2}^x f_1^x(v_{1k}^x)+b_{2}^x\\
    \vdots& &\vdots\\
    v_{L_xk}^x&=&A_{L_x} f_{L_x-1}^x(v_{(L_x-1)k}^x)+b_{L_x}^x\\
    x_{k+1}&=&v_{L_xk}^x
\ea\right.
\label{eq:RNN-x}
\end{equation}
\begin{equation}
\left\{\ba{rcl}
    v_{1k}^y&=&A_1^y\smallmat{x_k\\u_k}+b_1^y\\
    v_{2k}^y&=&A_{2}^y f_1^y(v_{1k}^y)+b_{2}^y\\
    \vdots& &\vdots\\
    v_{L_yk}^y&=&A_{L_y}^y f_{L_y-1}^y(v_{(L_y-1)k}^y)+b_{L_y}^y \\
    \hat y_k&=&f_{L_y}^y(v_{L_yk}^y)
\ea\right.
\label{eq:RNN-y}
\end{equation}
\label{eq:RNN-xy}%
\end{subequations}
where $L_x-1\geq 0$ and $L_y-1\geq 0$ are the number of hidden layers of the state-update 
and output functions, respectively, $v_i^x\in\rr^{n_i^x}$, $i=1,\ldots L_x-1$,
$v_{L_x}^x\in\rr^{n_x}$,
and $v_i^y\in\rr^{n_i^y}$, $i=1,\ldots L_y$, the values
of the corresponding inner layers,
$f_i^x:\rr^{n_i^x}\to \rr^{n_{i+1}^x}$, $i=1,\ldots L_x-1$,
$f_i^y:\rr^{n_i^y}\to \rr^{n_{i+1}^y}$, $i=1,\ldots L_y-1$
the corresponding activation functions, and $f_{L_y}^y:\rr^{n_{L_y}^y}\to \rr^{n_y}$
the output function, e.g., $f_{L_y}^y(v_{L_y}^y)=v_{L_y}^y$ to model
numerical outputs, or $[f_{L_y}^y(v_{L_y}^y)]_i=1/\left(1+e^{[v_{L_y}^y]_i}\right)$,
$i=1,\ldots,n_y$, for binary outputs. Note that the RNN~\eqref{eq:RNN} can be made
strictly causal by simply setting $v_{1}^y(k)=A_1^yx_k+b_1^y$ in~\eqref{eq:RNN-y}.

The training problem we want to solve is:
\begin{subequations}
\begin{eqnarray}
    \min_{z}&&r(x_0,\tx,\ty)+g(\tx,\ty)+\displaystyle{\sum_{k=0}^{N-1} \ell(y_{k},f_y(x_k,u_k,\ty))}\nonumber\\[-.6em]\label{eq:training-cost-reg}\\
    \mbox{s.t.}&& x_{k+1}=f_x(x_k,u_k,\tx),\ k=0,\ldots,N-2\label{eq:dyn-constraints-reg}
\end{eqnarray}
\label{eq:training}%
\end{subequations}
where $z=[x_0'\ \ldots\ x_{N-1}'\ \tx'\ \ty']'\in\rr^n$ is the overall optimization vector,
$n=Nn_x+n_{\tx}+n_{\ty}$,
$\ell:\rr^{n_y}\times\rr^{n_y}\to\rr$ is a loss function that we assume strongly convex and twice differentiable with respect to its second argument,
$r:\rr^{n_x}\times\rr^{n_{\tx}}\times\rr^{n_{\ty}}\to\rr$ is a regularization term (such as an $\ell_2$-regularizer) that we also assume strongly convex and twice differentiable,
and $g:\rr^{n_{\tx}}\times \rr^{n_{\tx}}\to\rr\cup\{+\infty\}$ is a 
possibly non-smooth and non-convex regularization term. For simplicity, in the following we assume that $r$ is separable with respect to $x_0$, $\tx$, and $\ty$, i.e., $r(x_0,\tx,\ty)=r_x(x_0)+r_\theta^x(\tx)
+r_\theta^y(\ty)$, and that in case of multiple outputs
$n_y>1$, the loss function $\ell$ is also separable, i.e.,
$\ell(y,\hat y)=\sum_{i=1}^{n_y}\ell_i(y_{ki},f_{yi}(x_k,u_k,\ty))$,
where the subscript $i$ denotes the $i$th component of the
output signal $y_k$ and the corresponding loss function.

Problem~\eqref{eq:training} can be rewritten as the following
unconstrained nonlinear programming problem in \emph{condensed form}
\begin{equation}
    \min_{x_0,\tx,\ty}V(x_0,\tx,\ty)+g(\tx,\ty)
    \label{eq:training-condensed}
\end{equation}
where we get\\[-1em]
\begin{equation}
    V(x_0,\tx,\ty)=r(x_0,\tx,\ty)+\sum_{k=0}^{N-1} \ell(y_{k},f_y(x_k,u_k,\ty))    
\label{eq:training-cost-condensed}
\end{equation}
by replacing $x_{k+1}=f_x(x_k,u_k,\tx)$ iteratively. Note that Problem~\eqref{eq:training-condensed} can be highly nonconvex and present several local minima,
see, e.g., recent studies reported in~\cite{RTUSA20}.

\subsection{Multiple training traces and initial-state encoder}
For simplicity of notation, in this paper we focus on training a RNN model~\eqref{eq:RNN-xy} based on a single input/output trace. The extension to $M$ multiple experiments is trivial and can be achieved by
introducing an initial state $x_{0,j}$ per trace and minimizing
the sum of the corresponding loss functions and the regularization terms with respect to
$(x_{0,1},\ldots,x_{0,M},\tx,\ty)$.  In alternative, we can parameterize
\begin{equation}
    x_{0,i}=f_{x0}(v_{0,i},\txo)
\label{eq:fx0}
\end{equation}
where $v_0\in\rr^{n_v}$ is a measured vector available at time 0, such as a collection of 
$n_a$ past outputs and $n_b$ past inputs as
suggested in~\cite{MB21,BTS21}, and/or other measurable values
that are known to affect the initial state of the system we want to model. The initial-state encoder function $f_{x0}:\rr^{n_v}\times\rr^{n_{\txo}}\to\rr^{n_x}$ is a feedforward neural network defined as in~\eqref{eq:RNN-xy}, parameterized by the vector $\txo\in\rr^{n_{\txo}}$ to be learned jointly with $\tx$, $\ty$. The regularization term in this case is defined as $r(\txo,\tx,\ty)$.

\section{Sequential linear least squares}
\label{sec:SLS}
We first handle the smooth case $g(\tx,\ty)=0$. Let $(x_0^0,\tx^0,\ty^0)$ be an initial guess. By
simulating the RNN model~\eqref{eq:RNN} we get the corresponding sequence of hidden states $x_k^0$, $k=1,\ldots,N-1$, and predicted outputs $\hat y_k^0$. At a generic iteration $h$ of the
algorithm proposed next, we assume that the nominal trajectory
is feasible, i.e., $x_{k+1}^h=f_x(x_k^h,u_k,\tx^h)$, $k=0,\ldots,N-2$. 
We want to find updates $x_k^{h+1}$, $\tx^{h+1}$, $\ty^{h+1}$ by 
solving a least-squares approximation of problem~\eqref{eq:training} as described in the next paragraphs.

\subsection{Linearization of the dynamic constraints}
To get a quadratic approximation of problem~\eqref{eq:training-condensed}
we first linearize the RNN dynamic constraints~\eqref{eq:dyn-constraints-reg} around $x_k^h$ by approximating $x_{k+1}^h+p_{x_{k+1}}$ $=$ $f_x(x_k^h+p_{x_k},u_k,\tx^h+p_{\tx})$ $\approx$ $f_x(x_k^h,u_k,\tx^h)$ $+$ $\nabla_x f_x(x_k^h,u_k,\tx^h)p_{x_k}$ $+$
    $\nabla_{\tx} f_x(x_k^h,u_k,\tx^h)p_{\tx}$.
Since $x_{k+1}^h=f_x(x_k^h,u_k,\tx^h)$, we can write the linearized state-update
equations as
\begin{equation}
    p_{x_{k+1}}=\nabla_x f_x(x_k^h,u_k,\tx^h)p_{x_k}+
    \nabla_{\tx} f_x(x_k^h,u_k,\tx^h)p_{\tx}
\label{eq:state-update-lin}
\end{equation}

The dynamics~\eqref{eq:state-update-lin} can be condensed to
\begin{equation}
    p_{x_{k}}=M_{kx}p_{x_0}+M_{k\tx}p_{\tx}
\label{eq:state-update-lin-cond}
\end{equation}
where matrices $M_{kx}\in\rr^{n_x\times n_x}$ and $M_{k\tx}\in\rr^{n_x\times n_{\tx}}$,
$k=0,\ldots,N-2$, are recursively defined as follows:
\beqar
    M_{0x}&=&I,\quad M_{0\tx}=0 \label{eq:M-updates}\\
    M_{(k+1)x}&=&\nabla_x f_x(x_k^h,u_k,\tx^h)M_{kx}\nonumber\\
        M_{(k+1)\tx}&=&\nabla_{x} f_x(x_k^h,u_k,\tx^h)M_{k\tx}
+\nabla_{\tx} f_x(x_k^h,u_k,\tx^h)\nonumber
\eeqar
In case $x_0$ is parameterized as in~\eqref{eq:fx0}, 
by approximating 
\[
x_0^h\approx f_{x0}(v_{0},\txo^h)+
\nabla_{\txo}f_{x0}(v_0,\txo^h)p_{\txo}
\]
where $p_{\txo}$ is the increment of $\txo$ to be computed,
it is enough to replace $p_{x_0}$ with
$p_{\txo}$ in~\eqref{eq:state-update-lin-cond} and $M_{0x}=\nabla_{\txo}f_{x0}(v_0,\txo^h)$ in~\eqref{eq:M-updates}.

\subsection{Quadratic approximation of the cost function}
Let us now find a quadratic approximation of the cost function~\eqref{eq:training-cost-condensed}.
First, regarding the regularization term $r$, we take the 2$^{\rm nd}$-order Taylor expansions
\beqarno
    r_x(x_0+p_{x_0})&\approx& \frac{1}{2}p_{x_0}'\nabla^2 r_x(x_0^h)p_{x_0}
+\nabla r_x(x_0^h)p_{x_0}+ r(x_0^h)\\
    r_\theta^x(\tx^h+p_{\tx})&\approx& \frac{1}{2}p_{\tx}'\nabla^2 r_\theta^x(\tx^h)p_{\tx}
+\nabla r_\theta^x(\tx^h)p_{\tx}+ r_\theta^x(\tx^h)\\
    r_\theta^y(\ty^h+p_{\ty})&\approx& \frac{1}{2}p_{\ty}'\nabla^2 r_\theta^y(\ty^h)p_{\ty}
+\nabla r_\theta^y(\ty^h)p_{\ty}+ r_\theta^y(\ty^h)
\eeqarno
By neglecting the constant terms, minimizing the 2nd-order Taylor expansion of $r(x_0+p_{x_0},\tx+p_{\tx},\ty+p_{\ty})$ is equivalent to minimizing
the least-squares term
\begin{equation}
    \frac{1}{2}\left\|\smallmat{L_{r0}p_{x0}+(L_{r0}')^{-1}\nabla_{x}r_x(x_0^h)\\
L_{rx}p_{\tx}+(L_{rx}')^{-1}\nabla r_\theta^x(\tx^h)\\
L_{ry}p_{\ty}+(L_{ry}')^{-1}\nabla r_\theta^y(\ty^h)
}\right\|_2^2   
\label{eq:LS-loss}
\end{equation}
where $L_{r0}'L_{r0}=\nabla^2r_x(x_0)$, $L_{rx}'L_{rx}=\nabla^2r_{\theta}^x(\tx)$,
and $L_{ry}'L_{ry}=\nabla^2r_{\theta}^y(\ty)$
are Cholesky factorizations of the corresponding Hessian matrices. Note that in the case of standard $\ell_2$-regularization
$r_x(x_0)=\frac{\rho_x}{2}\|x_0\|_2^2$,
$r_\theta^x(\tx)=\frac{\rho_\theta}{2}\|\tx\|_2^2$,
$r_\theta^y(\ty)=\frac{\rho_\theta}{2}\|\ty\|_2^2$,
we simply have $L_{r0}=\sqrt{\rho_x} I$, 
$L_{rx}=\sqrt{\rho_\theta} I$, $L_{ry}=\sqrt{\rho_\theta} I$, $\nabla_{x}r_x(x_0)=\sqrt{\rho_x} x_0$,
$r_\theta^x(\tx)=\sqrt{\rho_\theta}\tx$,
$r_\theta^y(\ty)=\sqrt{\rho_\theta} \ty$,
and hence~\eqref{eq:LS-loss} becomes
\begin{equation}
    \frac{1}{2}\left\|\smallmat{\sqrt{\rho_x}p_{x0}+\sqrt{\rho_x}x_0^h\\
\sqrt{\rho_\theta}p_{\tx}+\sqrt{\rho_\theta}\tx^h\\
\sqrt{\rho_\theta}p_{\ty}+\sqrt{\rho_\theta}\ty^h
}\right\|_2^2   
\label{eq:LS-loss-l2}
\end{equation}
Regarding the loss terms penalizing the $i$th output-prediction error
at step $k$, we have that
\beqarno
    \nabla_{x_k}\ell_i(y_{ki},f_{yi}(x_k^h,u_k,\ty^h))&=&\nabla_{x_k} f_{yi}(x_k^h,u_k,\ty^h)\ell'_{ik}\\
    \nabla_{\ty}\ell_i(y_{ki},f_{yi}(x_k^h,u_k,\ty^h))&=&\nabla_{\ty} f_{yi}(x_k^h,u_k,\ty^h)\ell'_{ik}
\eeqarno
where $\ell'_{ik}\eqdef\frac{d\ell_i(y_k,f_{yi}(x_k^h,u_k,\ty^h))}{d\hat y_i}$,
$\ell'_{ik}\in\rr$,
and hence, by differentiating again with respect to $\smallmat{x_k\\\ty}$
and setting $\ell''_{ik}\eqdef\frac{d^2\ell_i(y_k,f_{yi}(x_k^h,u_k,\ty^h)}{d\hat y_i^2}$, $\ell''_{ik}\in\rr$, $\ell''_{ik}>0$, we get
\[
  \nabla^2_{\smallmat{x_k\\\ty}\smallmat{x_k\\\ty}}
\ell_i(y_{ki},f_{yi}(x_k^h,u_k,\ty^h))=
(\nabla_{\smallmat{x_k\\\ty}} f_{yi})(\nabla_{\smallmat{x_k\\\ty}} f_{yi})'\ell''_{ik}
+\nabla^2_{\smallmat{x_k\\\ty}\smallmat{x_k\\\ty}}f_{yi}(x_k^h,u_k,\ty^h)\ell'_{ik}
\]
(we have omitted the dependence of $\ell'$, $\ell''$ on
the iteration $h$ for simplicity). By neglecting the Hessian of the output function $f_{yi}$
(as it is common in Gauss-Newton methods),
the minimization of $\ell_i(y_{ki},f_{yi}(x_k^h+p_{x_k},u_k,\tx^h+p_{\tx}))$ can be
approximated by the minimization of 
\[
  \frac{1}{2}\smallmat{p_{x_k}\\p_{\tx}}'
\ell''_{ik}(\nabla_{\smallmat{x_k\\\ty}} f_{yi})(\nabla_{\smallmat{x_k\\\ty}} f_{yi})'
\smallmat{p_{x_k}\\p_{\ty}}+\ell'_{ik}\smallmat{\nabla_{x_k} f_{yi}\\
\nabla_{\ty} f_{yi}}'\smallmat{p_{x_k}\\p_{\ty}}
\]
By strong convexity of the loss function $\ell$, this can be further rewritten as the minimization of the least-squares term
\begin{equation}
\frac{1}{2}\left\|(\ell''_{ik})^{\frac{1}{2}}(\nabla_{\smallmat{x_k\\\ty}} f_{yi})'
\smallmat{p_{x_k}\\p_{\ty}} +
(\ell''_{ik})^{-\frac{1}{2}}\ell'_{ik}\right\|_2^2    
\label{eq:loss-xk}
\end{equation}
that, by exploiting~\eqref{eq:state-update-lin-cond}, is equivalent to
\beqar
&&\frac{1}{2}\left\|(\ell''_{ik})^{\frac{1}{2}}
\matrice{ccc}{(\nabla_{x_k}f_{yi})'M_{kx} & (\nabla_{x_k}f_{yi})'M_{k\tx}&(\nabla_{\ty}f_{yi})'}\smallmat{p_{x_k}\\p_{\tx}\\p_{\ty}} +
(\ell''_{ik})^{-\frac{1}{2}}\ell'_{ik}\right\|_2^2    
\label{eq:1st-only}
\eeqar

Finally, the minimization of the quadratic approximation of~\eqref{eq:training-cost-condensed} can be recast as the linear LS problem
\begin{equation}
    p^h=\arg\min_p\frac{1}{2}\|A^hp-b^h\|_2^2   
\label{eq:LS}
\end{equation}
where $p=\smallmat{p_{x_0}\\p_{\tx}\\p_{\ty}}$,
$A^h$ and $b^h$ are constructed by collecting the least-squares terms from~\eqref{eq:LS-loss-l2} and~\eqref{eq:1st-only}, 
$A^h\in\rr^{n_A\times n_p}$,
$b^h\in\rr^{n_A}$, $n_A=Nn_y+n_x+n_{\tx}+n_{\ty}$, and $n_p=n_x+n_{\tx}+n_{\ty}$.

We remark that most of the computation effort is usually spent
in computing the Jacobian matrices $\nabla_{\smallmat{x_k\\\tx}} f_{x}(x_k^h,u_k,\tx^h)$
and $\nabla_{\smallmat{x_k\\\ty}} f_{yi}(x_k^h,u_k,\ty^h)$. 
Given the current nominal trajectory $\{x_k^h\}$, such computations can
be completely \emph{parallelized}. Hence, we are not forced to use \emph{sequential} computations 
as in backpropagation~\cite{Wer90} to solve for $p^h$ in~\eqref{eq:LS}, as
typically done in classical training methods based on gradient-descent.

Note also that for training \emph{feedforward} neural networks we 
set $n_x=0$, $v_{1}^y(k)=A_1^yu_k+b_1^y$ in~\eqref{eq:RNN-y},
and only optimize with respect to $\ty$. In this case, 
the operations performed at each iteration $h$ to predict $\hat y_k$
and compute $V^h$ can be also parallelized.

For the sake of comparison, a gradient-descent method would set
\beqar
    p^h&=&-\nabla V(x_0^h,\tx^h,\ty^h) \label{eq:grad-V}\\
       &=&\smallmat{
    \nabla_{x_0}r_x(x_0^h)\\\nabla_{\tx}r_{\theta}^x(\tx^h)\\\nabla_{\ty}r_{\theta}^y(\ty^h)}%\\&&
    +
    \sum_{k=0}^{N-1}\sum_{i=1}^{n_y}\ell'_{ik}\smallmat{
    \nabla_{x_k}f_{yi}(x_k^h,\ty^h)'M_{kx}\\\nabla_{x_k}f_{yi}(x_k^h,\ty^h)'M_{k\tx}\\\nabla_{\ty}f_{yi}(x_k^h,\ty^h)
    } \nonumber
\eeqar
since, from~\eqref{eq:M-updates}, $\nabla_{x_0}x_k=M_{kx}$ and $\nabla_{\tx}x_k=M_{k\tx}$.

\subsubsection{Sequential unconstrained quadratic programming}
\label{sec:SQP}
The formulation~\eqref{eq:LS} requires storing matrix $A^h$ and vector $b^h$.
In order to decrease memory requirements, especially in the case of a large number $N$
of training data, an alternative method is to incrementally construct the matrices $H^h=(A^h)'A^h\in\rr^{n_p\times n_p}$
and $c^h=-(A^h)'b^h\in\rr^{n_p}$ of the normal equation  $H^hp^h+c^h=0$ associated 
with~\eqref{eq:LS}. Although this approach might be computationally
more expensive, it only requires storing $\frac{n_p^2+3n_p}{2}$ numbers.
A further advantage of solving the normal equation is that the
convexity of $\ell$ with respect to $\hat y$ and of $r(x,\tx,\ty)$ do not have to be strong, as instead required to express the quadratic approximation of $f$ as a squared norm
from~\eqref{eq:1st-only} and~\eqref{eq:LS-loss}, respectively, as long as the computed $H^h$ is nonsingular. 

\subsubsection{Recursive least-squares}
A further alternative to precomputing and storing the matrices $A^h,b^h$ is to
exploit~\eqref{eq:1st-only} and use the
following recursive least-square (RLS) formulation of~\eqref{eq:LS}
\beqar
    P_{k}&=&P_{k-1}-\frac{P_{k-1}\phi_k\phi_k'P_{k-1}}{1+\phi_k'P_{k-1}\phi_k}\nonumber\\
    p_k&=&p_{k-1}+P_k\phi_k(\eta_k-p_{k-1}'\phi_k)
    \label{eq:RLS}
\eeqar
for $k=0,\ldots,N-1$
where $\phi_k=\sqrt{\ell''_{ik}}\smallmat{M'_{kx}\nabla_{x_k} f_{yi}\\M'_{k\tx}\nabla_{x_k} f_{yi}\\
\nabla_{\ty} f_{yi}}$, $\eta_k=\ell'_{ik}/\sqrt{\ell''_{ik}}$.
The RLS iterations in~\eqref{eq:RLS} provide the optimal solution $p^h\eqdef p_{N-1}$ (as well as  $P_{N-1}=((A^h)'A^h)^{-1})$
when the matrix $P_k\in\rr^{n_p\times n_p}$ and vector $p_k$ are initialized,
respectively, as
\begin{equation}
    P_{-1}=\smallmat{\frac{1}{\rho_x}I_{n_x}&0&0\\
    0&\frac{1}{\rho_\theta}I_{n_{\tx}}&0\\0&0&\frac{1}{\rho_\theta}I_{n_{\tx}}},
    \quad
    p_{-1}=\smallmat{x_0^h\\\ty^h\\\tx^h}
    \label{eq:RLS-init}
\end{equation}
Different algorithms were proposed in the literature to implement RLS efficiently, 
such as the inverse QR method proposed in~\cite{AG93}, in which 
a triangular matrix $R_k$ is updated recursively starting from $R_{-1}=(P_{-1})^{\frac{1}{2}}$.

\subsection{Line search}
To enforce the decrease of the cost function~\eqref{eq:training-cost-reg}
across iterations $h=0,1,\ldots$, a possibility is to update the solution as
\begin{equation}
    \smallmat{
    x_0^{h+1}\\
    \tx^{h+1}\\
    \ty^{h+1}}=\smallmat{x_0^h\\\tx^h\\\ty^h}+\alpha_h\smallmat{p^h_{x_0}\\p^h_{\tx}\\p^h_{\ty}}
\label{eq:update}
\end{equation}
and set $x_{k+1}^{h+1}=f_x(x_k^{h+1},u_k,\tx^{h+1})$, $k=1,\ldots,N-1$. 
We choose the step-size $\alpha_h$ by finding the largest $\alpha$
satisfying the Armijo condition 
\begin{equation}
    V^{h+1}\leq V^h+c_1\alpha(\nabla V^h)'p^h 
\label{eq:Armijo}
\end{equation}
where $V^{h+1}$, $V^{h}$ are the costs defined as in~\eqref{eq:training}
associated with the corresponding combinations of $(x_0,\tx,\ty)$,
the gradient is taken with respect to $\smallmat{x_0\\\tx\\\ty}$,
and $c_1$ is a constant (e.g., $c_1=10^{-4}$)~\cite[Chapter 3.1]{NW06}. In this paper
we use the common approach of starting with $\alpha=1$ and then
change $\alpha\leftarrow \sigma\alpha$, $\sigma\in(0,1)$ until~\eqref{eq:Armijo}
is satisfied. We set a bound $N_{\sigma}$ on the number of line-search
steps that can be performed, setting $\alpha=0$ in the case of line-search failure.
Since $V(x_0+\alpha p_{x_0}^h,\tx+\alpha p_{\tx}^h,\ty+\alpha p_{\ty}^h)\approx 
V(x_0,\tx^h,\ty^h)+\frac{1}{2}\|A^h\alpha p^h-b^h\|_2^2-\frac{1}{2}\|b^h\|_2^2$
for small values of $\alpha$, the directional derivative in~\eqref{eq:Armijo} can be computed from $A^h,b^h$:
\[
    \ba{rcl}
    (\nabla V^h)'p^h&=&\underset{\alpha\rightarrow 0}{\lim}
\frac{V(x_0+\alpha p_{x_0}^h,\tx+\alpha p_{\tx}^h,\ty+\alpha p_{\ty}^h)-V(x_0,\tx^h,\ty^h)}{\alpha}\\
&=&\underset{\alpha\rightarrow 0}{\lim}\frac{1}{2}\frac{\|A^h\alpha p^h-b^h\|_2^2-\|b^h\|_2^2}{\alpha}
=-(b^h)'A^hp^h
\ea
\]
Note that when the loss function $\ell(y,\hat y)=\frac{1}{2}\|y-\hat y\|_2^2$ and the regularization terms $r_x,r_{\ty},r_{\tx}$ are quadratic, the loss 
$V(x_0,\tx^h,\ty^h)=\frac{1}{2}(b^h)'b^h$ by construction.

The overall nonlinear least-squares solution method 
for training recurrent neural networks described in the previous sections is summarized in Algorithm~\ref{algo:SLS}.

Algorithm~\ref{algo:SLS} belongs to the class of Generalized Gauss-Newton (GGN) methods~\cite{MBD21} applied to solve problem~\eqref{eq:training-cost-condensed},
with the addition of line search. In fact, following the notation in~\cite{MBD21},
we can write $V(x_0,\tx,\ty)=\Phi_0(F(w))$ where $w\eqdef[x_0'\ {\tx}' \ {\ty}]'$
is the optimization vector, function $\Phi_0:\rr^{N_\phi}\to\rr$,
$N_\phi\eqdef n_x+n_{\tx}+n_{\ty}+Nn_y$, is given by
the composition $\Phi_0=\Phi_+\circ\Phi_\ell$ of the summation function 
$\Phi_+:\rr^{N_\phi}\to\rr$, $\Phi_+(x)=\sum_{i=1}^{N_\phi}x_i$, 
with the vectorized loss $\Phi_\ell:\rr^{N_\phi}\to\rr^{N_\phi}$,
$\Phi_\ell([x_0'\ \tx'\ \ty'\ \hat y_0'\ \ldots\ \hat y_{N-1}']')=[r_x(x_0)\ r_\theta^x(\tx)\ r_\theta^y(\ty)\ \ell(y_0,\hat y_0)'\ \ldots
\ell(y_{N-1},\hat y_{N-1})']'$. By assumption, all the components of $\Phi_\ell$
are strongly convex functions, and since $\Phi_+$ is also strongly convex,
$\Phi_0$ is strongly convex with
respect to its argument $F(w)$. As in GGN methods, Algorithm~\ref{algo:SLS}
takes a second-order approximation of $\Phi_0$ and a linearization
of $F(w)$ during the iterations. In~\cite{MBD21}, local convergence under the full
step $\alpha_h\equiv 1$, $\forall h\geq 0$, is shown under additional assumptions
on $\Phi_0$ and the neglected second-order derivatives of $F$, while in Algorithm~\ref{algo:SLS} we force monotonic decrease of the loss function $V$ at each iteration $h$ by line search, hence obtaining a ``damped'' GGN algorithm. Note that the latter may fail ($\alpha_h=0$) in the case $p^h$ is not a descent direction or $N_{\sigma}$ is not large enough. In such a case, Algorithm~\ref{algo:SLS} would stop immediately as $V^{h}-V^{h-1}=0$.

\begin{algorithm}[t]
    \caption{Sequential least squares with line search.}
    \label{algo:SLS}
    ~~\textbf{Input}: Training dataset $\{u_k,y_k\}_{k=0}^{N-1}$, initial guess $(x_0^0$, $\tx^0$, $\ty^0)$, maximum number
$E$ of epochs, tolerance $\epsilon_V>0$, line-search parameters $c_1>0$, $\sigma\in(0,1)$,
$N_\sigma\geq 1$.
    \vspace*{.1cm}\hrule\vspace*{.1cm}
    \begin{enumerate}[label*=\arabic*., ref=\theenumi{}]
        \item Simulate~\eqref{eq:RNN} to get $x_{k+1}^0$, $k=0,\ldots,N-1$,
              and the initial cost $V^0$ from~\eqref{eq:training-cost-condensed};\label{step:init-1}
        \item $h\leftarrow 0$;\label{step:init-2}
        \item \textbf{repeat}:
        \begin{enumerate}[label=\theenumi{}.\arabic*., ref=\theenumi{}.\arabic*]
        \item Construct and solve the linear least-squares problem~\eqref{eq:LS}
        to get a descent direction $p^h$;
        \item Find step-length $\alpha_h\in[0,1]$ by line-search using Armijo 
        rule~\eqref{eq:Armijo} and get the new cost $V^{h+1}$;
        \item Update $x_0^{h+1},\tx^{h+1},\ty^{h+1}$ as in~\eqref{eq:update};
        \item $h\leftarrow h+1$;
        \end{enumerate}
        \item \textbf{until} $V^{h}-V^{h-1}\leq \epsilon_V$ or $h\geq E$;
        \item \textbf{end}.
    \end{enumerate}
    \vspace*{.1cm}\hrule\vspace*{.1cm}
    ~~\textbf{Output}: RNN parameters $\tx^{h},\ty^{h}$ and initial hidden state $x_0^{h}$.
\end{algorithm}

\subsection{Initialization}
\label{sec:initialization}
In this paper we initialize $x_0^0=0$, set zero initial bias terms, and draw the remaining
components of $\tx^0$ and $\ty^0$ from the normal distribution with zero mean and standard deviation defined as in~\cite{GB10}, further multiplied by the quantity $\sigma_0\leq1$.
The rationale to select small initial values of $\tx^0$, $\ty^0$ is to avoid that
the initial dynamics $f_x$ used to compute $x_k^0$ for $k=1,\ldots,N-1$
are unstable. Then, since the overall loss $V^h$ is decreasing, the divergence of
state-trajectories is unlikely to occur at subsequent iterations.
Note that this makes the elimination of the state variables $x_1,\ldots,x_{N-1}$
in the condensing approach~\eqref{eq:training-condensed}
suitable for the current RNN training setting, contrarily to solving 
nonlinear MPC problems in which the initial state $x_0$ and the model
coefficients $\tx$, $\ty$ are fixed and may therefore lead to excite
unstable linearized model responses, with consequent numerical issues~\cite{BC22}.
Note also that ideas to enforce open-loop stability could be introduced in our setting
by adopting ideas as in~\cite{KM19}, where the authors propose to also learn a Lyapunov function while learning the model.

\subsection{A Levenberg-Marquardt variant}
Trust-region methods are an alternative family of approaches to line-search 
for forcing monotonicity of the objective function. In solving nonlinear
least-squares problems, the well-known Levenberg-Marquardt (LM) algorithm~\cite{Lev44,Mar63}
is often used due to its good performance. In LM methods,~\eqref{eq:LS} is changed
to
\begin{equation}
    p^h=\arg\min_p\frac{1}{2}\|A^hp-b^h\|_2^2+\frac{1}{2}\lambda_h\|p\|_2^2
    \label{eq:LM}
\end{equation}
where the regularization term $\lambda_h$ is tuned at each iteration $h$ 
to meet the condition $V^h\leq V^{h-1}$. It is easy to show
that tuning $\lambda_h$ is equivalent to tuning the trust-region radius, see e.g.~\cite[Chapter 10]{NW06}. Clearly, setting $\lambda_h=0$ corresponds
to the full Generalized Gauss-Newton step, as in line-search when $\alpha_h=1$,
and $\lambda_h\rightarrow\infty$ corresponds to $p^h\rightarrow 0$ ($\alpha_h\rightarrow 0$).
Several methods exist to update $\lambda_h$ (and possibly generalizing $\|p\|_2^2$
to $\|D^hp\|_2^2$ for some weight matrix $D^h$)~\cite{TS12}. In this paper, we take the simplest approach suggested in~\cite[Section 2.1]{TS12} and, starting from $\lambda_h=\lambda_{h-1}$, keep multiplying $\lambda_h\leftarrow c_2\lambda_h$ until the condition $V^h\leq V^{h-1}$ is met, then reduce $\lambda_h\leftarrow \lambda_h/c_3$ before proceeding
to step $h+1$. The approach is summarized in Algorithm~\ref{algo:LM}.

\begin{algorithm}[t]
    \caption{Sequential least squares, Levenberg-Marquardt variant.}
    \label{algo:LM}
    ~~\textbf{Input}: Training dataset $\{u_k,y_k\}_{k=0}^{N-1}$, initial guess $(x_0^0$, $\tx^0$, $\ty^0)$, maximum number
$E$ of epochs, tolerance $\epsilon_V>0$, initial value $\lambda_0>0$, LM parameters $c_2,c_3>1$,
number $N_\lambda\geq 1$ of search iterations.
    \vspace*{.1cm}\hrule\vspace*{.1cm}
    \begin{enumerate}[label*=\arabic*., ref=\theenumi{}]
        \item Simulate~\eqref{eq:RNN}, get $x_{k+1}^0$, $k=0,\ldots,N-1$,
              and initial cost $V^0$ from~\eqref{eq:training-cost-condensed};\label{step:init-1}
        \item $h\leftarrow 0$;\label{step:init-2}
        \item \textbf{repeat}:
        \begin{enumerate}[label=\theenumi{}.\arabic*., ref=\theenumi{}.\arabic*]
            \item Construct the least-squares problem~\eqref{eq:LS};
            \item $j\leftarrow 0$;
            \item \textbf{repeat}:
            \begin{enumerate}[label=\theenumii{}.\arabic*., ref=\theenumii{}.\arabic*]
                \item Solve the regularized linear least-squares problem~\eqref{eq:LM}
                and get a candidate $p^h$;
                \item Update $x_0^{h+1},\tx^{h+1},\ty^{h+1}$ as in~\eqref{eq:update} when $\alpha_h=1$;
                \item Get new cost $V^{h+1}$;
                \item \textbf{if} $V^{h+1}\leq V^h$ decrease $\lambda_h\leftarrow\lambda_h/c_3$;
                \item \textbf{else} increase $\lambda_h\leftarrow c_2\lambda_h$;
                \item \textbf{set} $j\leftarrow j+1$;
            \end{enumerate}
            \item \textbf{until} $V^{h+1}\leq V^{h}$ or $j=N_\lambda$;
            \item $h\leftarrow h+1$;
            \item \textbf{if} $V^{h}>V^{h-1}$ \textbf{go to} Step~\ref{algo:LM:end};\label{algo:LM:failure}
        \end{enumerate}
        \item \textbf{until} $V^{h}-V^{h-1}\leq \epsilon_V$, or $h\geq E$;
        \item \textbf{end}.\label{algo:LM:end}
    \end{enumerate}
    \vspace*{.1cm}\hrule\vspace*{.1cm}
    ~~\textbf{Output}: RNN parameters $\tx^{h},\ty^{h}$ and initial hidden state $x_0^{h}$.
\end{algorithm}

Similarly to Algorithm~\ref{algo:SLS}, in Algorithm~\ref{algo:LM}
we have set a bound $N_{\lambda}$ on the number of least-squares
problems that can be solved at each epoch $h$. In the case an improvement
$V^h\leq V^{h-1}$ has not been found, Algorithm~\ref{algo:LM}
stops due to Step~\ref{algo:LM:failure}.

The main difference between performing line-search and computing instead LM
iterations is that at each epoch $h$ the former only solves the linear
least-squares problems~\eqref{eq:LS}, the latter requires solving up to $N_\lambda$
problems~\eqref{eq:LM}. Note that, however, both only construct 
$A^h$ and $b^h$ once at each iteration $h$, which, as remarked earlier, is usually the dominant
computation effort due to the evaluation of the required Jacobian matrices
$\nabla_xf_x$, $\nabla_{\tx}f_x$, $\nabla_xf_y$, $\nabla_{\ty}f_y$.

\section{Non-smooth regularization}
\label{sec:ADMM}
We now want to solve the full problem~\eqref{eq:training}
by considering also the additional non-smooth regularization term $g(\tx,\ty)$.
Consider again the condensed loss function~\eqref{eq:training-cost-condensed} and rewrite problem~\eqref{eq:training} according to the following split
\beqar
    \min_{x_0,\tx,\ty,\nu_x,\nu_y} &&V(x_0,\tx,\ty)+g(\nu_x,\nu_y)\nonumber\\
    \st&&\tx=\nu_x,\quad \ty=\nu_y
    \label{eq:training-reg-split}%
\eeqar
We solve problem~\eqref{eq:training-reg-split} by executing
the following scaled ADMM iterations
\begin{subequations}
\beqar
    \smallmat{x_0(t+1)\\\tx(t+1)\\\ty(t+1)}&=&\arg\hspace*{-.7em}\min_{x_0,\tx,\ty}
    \hspace*{-.3em}
V(x_0,\tx,\ty)+\frac{\rho}{2}\left\|
    \hspace*{-.2em}\smallmat{\tx-\nu_x(t)+w_x(t)\\\ty-\nu_y(t)+w_y(t)}\hspace*{-.2em}\right\|_2^2\nonumber\\\label{eq:ADMM-x}\\
\smallmat{\nu_x(t+1)\\\nu_y(t+1)}&=&\arg\hspace*{-.2em}\min_{\nu_x,\nu_y}
\frac{1}{\rho}g(\nu_x,\nu_y)+\frac{1}{2}\left\|\hspace*{-.2em}\smallmat{\nu_x-\tx(t+1)-w_x(t)\\
\nu_y-\ty(t+1)-w_y(t)}\hspace*{-.2em}\right\|_2^2
\nonumber\\&=&\prox_{\frac{1}{\rho}g}(
\tx(t+1)+w_x(t),\ty(t+1)+w_y(t))\label{eq:ADMM-nu}\\
\smallmat{w_x(t+1)\\w_y(t+1)}&=&\smallmat{w_x^{h}+\tx(t+1)-\nu_x(t+1)\\w_y^{h}+\ty(t+1)-\nu_y(t+1)}\label{eq:ADMM-w}
\eeqar
\label{eq:ADMM}%
\end{subequations}
for $t=0,\ldots,N_{\rm ADMM}-1$, where ``$\prox$'' denotes the proximal operator. Problem~\eqref{eq:ADMM-x} is solved
by running either Algorithm~\ref{algo:SLS} or~\ref{algo:LM} with initial condition
$x_0^0=x_0(t)$, $\tx^0=\tx(t)$, $\ty^0=\ty(t)$. Note that when solving the least-squares problem~\eqref{eq:LS} or~\eqref{eq:LM}, $A^h,b^h$ are augmented to include the additional $\ell_2$-regularization term $\frac{1}{2}\left\|\sqrt{\rho}\smallmat{p_{\tx}\\p_{\ty}}
    -\sqrt{\rho}\smallmat{\nu_x(t)-\tx^h-w_x(t)\\\nu_y(t)-\ty^h-w_y(t)}\right\|_2^2$
introduced in the ADMM iteration~\eqref{eq:ADMM-x}. Note also that for $t>0$ 
Step~\ref{step:init-1} of Algorithms~\ref{algo:SLS} and~\ref{algo:LM} can be omitted,
since the initial simulated trajectory $x_{k+1}^0$ is available from the last execution of
the ADMM step~\eqref{eq:ADMM-x}, $\forall k=0,\ldots,N-2$,
and $V_0$ can be obtained by adjusting the cost computed at the last iteration
of the previous execution of Algorithm~\ref{algo:SLS} or~\ref{algo:LM}
to take into account that $\nu_x$, $\nu_y$, $w_x$, $w_y$ have possibly changed
their value.

Our numerical experiments show
that setting $E=1$ is a good choice, which corresponds
to just computing one sequential LS iteration to solve~\eqref{eq:ADMM-x} approximately.
The overall algorithm, that we call NAILS (nonconvex ADMM iterations and least squares
with line search) in the case Algorithm~\ref{algo:SLS} is used, or
NAILM (nonconvex ADMM iterations and least squares with Levenberg-Marquardt variant)
when instead Algorithm~\ref{algo:LM} is used, is summarized in 
Algorithm~\ref{algo:NAILS}. Note that NAILS/NAILM
processes the training dataset for a maximum of $E\cdot N_{\rm ADMM}$ epochs.

\begin{algorithm}
    \caption{Nonconvex ADMM iterations and least squares,
    either with a line-search (NAILS) or Levenberg-Marquardt (NAILM) approach}
    \label{algo:NAILS}
    ~~\textbf{Input}: Training dataset $\{u_k,y_k\}_{k=0}^{N-1}$, initial guess $x_0^0$, $\tx^0$, $\ty^0$, number $N_{\rm ADMM}$ of ADMM iterations, ADMM parameter $\rho>0$, maximum number $E$ of epochs, tolerance $\epsilon_V>0$; line-search parameters $c_1>0$, $\sigma\in(0,1)$, and $N_\sigma>1$ (NAILS), or LM parameters $\lambda_0>0$, $c_2,c_3>1$, and $N_\lambda>0$ (NAILM).
    \vspace*{.1cm}\hrule\vspace*{.1cm}
    \begin{enumerate}[label*=\arabic*., ref=\theenumi{}]
        \item $\smallmat{\nu_x^{0}\\\nu_y^{0}}\leftarrow\smallmat{\tx^0\\\ty^0}$, $\smallmat{w_x^{0}\\w_y^{0}}\leftarrow0$;
        \item \textbf{for} $t=0,1,\ldots,N_{\rm ADMM}-1$ \textbf{do}: 
        \begin{enumerate}[label=\theenumi{}.\arabic*., ref=\theenumi{}.\arabic*]
        \item \textbf{run} Algorithm~\ref{algo:SLS} or~\ref{algo:LM}
        to update $x_0(t+1)$, $\tx(t+1)$, and $\ty(t+1)$ as in~\eqref{eq:ADMM-x};
        \item \textbf{update} $\smallmat{\nu_x(t+1)\\\nu_y(t+1)}$ as in~\eqref{eq:ADMM-nu}
and $\smallmat{w_x(t+1)\\w_y(t+1)}$ as in~\eqref{eq:ADMM-w};
        \item $t\leftarrow t+1$;
        \end{enumerate}
        \item \textbf{end}.
    \end{enumerate}
    \vspace*{.1cm}\hrule\vspace*{.1cm}
    ~~\textbf{Output}: RNN parameters $\tx=\nu_x(t),\ty=
\nu_y(t)$ and initial hidden state $x_0(t)$.
\end{algorithm}

In NAILS, if RLS are used in alternative to forming and solving~\eqref{eq:LS},
since 
\[
  \arg\min_{\alpha\in\rr}\frac{1}{2}(\rho_1(\alpha-\alpha_1)^2+\rho_2(\alpha-\alpha_2)^2)= \arg\min_{\alpha\in\rr}\frac{\rho_1+\rho_2}{2}\left(\alpha-\frac{\rho_1\alpha_1+\rho_2\alpha_2}{\rho_1+\rho_2}\right)^2
\]
for any $\rho_1,\rho_2>0$ and $\alpha_1,\alpha_2\in\rr$, 
we can initialize
\[
    P_{-1}\hspace*{-.3em}=\hspace*{-.3em}\smallmat{\hspace*{-.1cm}\frac{1}{\rho_x}I_{n_x}&0&0\\
    0&\hspace*{-.3cm}\frac{1}{\rho_\theta+\rho}I_{n_{\tx}}&0\\0&0&\hspace*{-.3cm}\frac{1}{\rho_\theta+\rho}
    I_{n_{\ty}}\hspace*{-.05cm}}\hspace*{-.1cm},\
    p_{-1}\hspace*{-.3em}=\hspace*{-.3em}\smallmat{\hspace*{-.1cm}
    x_0^h\\
    \frac{\rho_\theta \tx^h+\rho (\nu_x(t)-\tx^h-w_x(t))}{\rho_\theta+\rho}\\
    \frac{\rho_\theta \ty^h+\rho (\nu_y(t)-\ty^h-w_y(t))}{\rho_\theta+\rho}
    \hspace*{-.1cm}}
\]
The final model parameters are given by $\tx=\nu_x(t)$,
$\ty=\nu_y(t)$, $x_0(t)$, where $t=N_{\rm ADMM}$ is the last ADMM iteration
executed. In case $x_0$
also needs to be regularized by a non-smooth penalty, we can extend
the approach to include a further split $\nu_{x_0}=x_0$.

Algorithm~\ref{algo:NAILS} can be modified to return
the initial state and model obtained after each ADMM step %(so to include the projection step)
that provide the best loss $\sum\ell(y_k,\hat y_k)$ 
observed during the execution of the algorithm
on training data or, if available, on a separate validation dataset. 

Note that the iterations~\eqref{eq:ADMM}
are not guaranteed to converge to a global optimum of the problem.
Moreover, it is well known that in some 
cases nonconvex ADMM iterations may even diverge~\cite{HFD16}. The reader
is referred to~\cite{HFD16,HLR16,WYZ19,TP20} for convergence conditions
and/or alternative ADMM formulations.

We finally remark that, in the absence of the non-smooth regularization
term $g$, Algorithm~\ref{algo:NAILS} simply reduces
to a smooth GNN method running with tolerance $\epsilon_V$ for a maximum
number $E$ of epochs.

\subsection{$\ell_1$-regularization}
A typical instance of regularization is the $\ell_1$-penalty
\begin{equation}
    g(\tx,\ty)=\tau_x\|\tx\|_1+\tau_y\|\ty\|_1
\label{eq:L1-reg-fun}
\end{equation}
to attempt pruning an overly-parameterized network structure, with $\tau_x,\tau_y\geq0$.
In this case, the update in~\eqref{eq:ADMM-nu}
simply becomes
\begin{equation}\ba{rcl}
    \nu_x(t+1)&=&S_{\frac{\tau_x}{\rho}}(\tx(t+1)+w_x^h)\\
    \nu_y(t+1)&=&S_{\frac{\tau_y}{\rho}}(\ty(t+1)+w_y^h)
    \ea
\label{eq:soft-threshold}%
\end{equation}
where $S_{\alpha}$ is the soft-threshold operator
\[
    [S_\alpha(w)]_i=\left\{\ba{rll}
    w_i+\alpha&\mbox{if}&w_i\leq -\alpha\\
   0&\mbox{if}&|w_i|\leq\alpha\\
    w_i-\alpha&\mbox{if}&w_i\geq \alpha
\ea\right.
\]

\subsection{Group-Lasso regularization and model reduction}
\label{sec:groupLasso}
The complexity of the structure
of the networks $f_x$, $f_y$ can be reduced by using group-Lasso
penalties~\cite{YL06} that attempt zeroing entire subsets of parameters. 
In particular,
the order $n_x$ of the RNN can be penalized 
by creating $n_x$ groups $\theta_i^g$ of variables, 
$\theta_i^g\in\rr^{n_g}$, $n_{g}=n_1^x+n_1^y+n_{L_x-1}^x+1$,
$i=1,\ldots,n_x$,
each one stacking the $i$th columns of $A_1^x$ and $A_1^y$,
the $i$th row of $A_{L_x}$, and the $i$th entry of $b_{L_x}$.
In this case, in~\eqref{eq:training-reg-split}
we only introduce the splitting $\theta_i^g=\nu_i^g$ on those variables and penalize
\[
    g(\nu_i^g)=\tau_g\sum_{i=1}^{n_x}\|\nu_i^g\|_2
\]
where $\tau_g\geq 0$ is the group-Lasso penalty.
Then, vectors $\nu_i^{g}(t+1)$ are updated
as in~\eqref{eq:soft-threshold} by replacing $S_{\frac{\tau_g}{\rho}}$
with the block soft thresholding operator $S_\frac{\tau_g}{\rho}^g:\rr^{n_{g}}\to\rr^{n_{g}}$
defined as
\[
    S_\alpha^g(w)=\left\{\ba{rll}
    (1-\frac{\alpha}{\|w\|_2})w&\mbox{if}&\|w\|_2> \alpha\\
   0&\mbox{if}&\|w\|_2\leq\alpha
\ea\right.
\]
The larger the penalty $\tau_g$, the larger is expected the number
of state indices $i$ such that the $i$th column of $A_1^x$ and $A_1^y$,
the $i$th row of $A_{L_x}$, and the $i$th entry of $b_{L_x}$ are all zero. 
As a consequence, the corresponding $i$th hidden state has no influence in the model.

\subsection{$\ell_0$-regularization}
The $\ell_0$-regularization term
\begin{equation}
    g(\tx,\ty)=\tau_x\|\tx\|_0+\tau_y\|\ty\|_0
\label{eq:L0-reg-fun}
\end{equation}
with $\tau_x^0,\tau_y^0\geq 0$ 
also admits the explicit proximal ``hard-thresholding'' operator 
\[
    [\prox_{\frac{\tau}{\rho}\|x\|_0}]_i=\left\{\ba{rll}
        0 & \mbox{if} & x_i^2<\frac{2\rho}{\tau}\\
        x_i & \mbox{if} & x_i^2\geq \frac{2\rho}{\tau}
    \ea\right.
\]
and can be used in alternative to~\eqref{eq:L1-reg-fun}.

\subsection{Integer constraints}
Constraints on $\tx$, $\ty$ that can expressed as
mixed-integer linear inequalities
\begin{equation}
    A_{\rm MI}\smallmat{\tx\\\ty}\leq b_{\rm MI},\quad
     {\tx}_{i},{\ty}_{j}\in\{0,1\},\ \forall i\in I_x,\ j\in I_y    
\label{eq:MI-constraints}
\end{equation}
where $I_x\subseteq\{1,\ldots,n_{\tx}\}$, $I_y\subseteq\{1,\ldots,n_{\ty}\}$,
$A_{\rm MI}\in\rr^{n_{\rm MI}\times (n_{\tx}+n_{\ty})}$,
$b_{\rm MI}\in\rr^{n_{\rm MI}}$, $n_{\rm MI}\geq 0$,
can be handled by setting $g$ as the indicator function
\[
    g(\tx,\ty)=\left\{\ba{rl}
    0 &\mbox{if~\eqref{eq:MI-constraints} is satisfied}\\
    +\infty&\mbox{otherwise}
\ea\right.
\]
In this case, $\nu_x^{h+1},\nu_y^{h+1}$ can be computed from~\eqref{eq:ADMM-nu} by solving the mixed-integer quadratic programming (MIQP) problem
\begin{equation}
\ba{rl}
\min_{\nu_x,\nu_y}&\frac{1}{2}\left\|\smallmat{\nu_x-\tx(t+1)-w_x(t)\\
\nu_y-\ty(t+1)-w_y(t)}\right\|_2^2\\
\st&A_{\rm MI}\smallmat{\tx\\\ty}\leq b_{\rm MI}\\
&{\tx}_{i},{\ty}_{j}\in\{0,1\}\ \forall i\in I_x,\ j\in I_y
\ea
\label{eq:MIQP-ADMM}
\end{equation}
Note that in the special case $n_{\rm MI}=0$,~\eqref{eq:MIQP-ADMM}
has the following explicit solution~\cite{TMBB17}
\begin{equation}
    \ba{rcl}
    \nu_{xi}(t+1)&=&\round({\tx}_i(t+1)+w_{xi}(t)),\ i\in I_x\\
    \nu_{yj}(t+1)&=&\round({\tx}_j(t+1)+w_{xj}(t)),\ j\in I_y
    \ea    
\label{eq:binary_quantization}
\end{equation}
where $\round(\alpha)=0$ if $\alpha<0.5$, or $1$ otherwise.

Binary constraints can be extended to more general quantization constraints
${\tx}_i,{\ty}_j\in \QQ\eqdef\{q_1,\ldots,q_{n_Q}\}\subset\rr$, $\forall i\in I_x$, 
$\forall j\in I_y$.
In fact, these can be handled as in~\eqref{eq:binary_quantization}
by setting $\nu_{xi}(t+1)$ equal to the value in $\QQ$ %$V_{xi}$ 
that is closest to ${\tx}_i(t+1)+w_{xi}^h$, and similarly for $\nu_{yj}(t+1)$.

\begin{figure}[t]
\centerline{
\includegraphics[width=0.8\hsize]{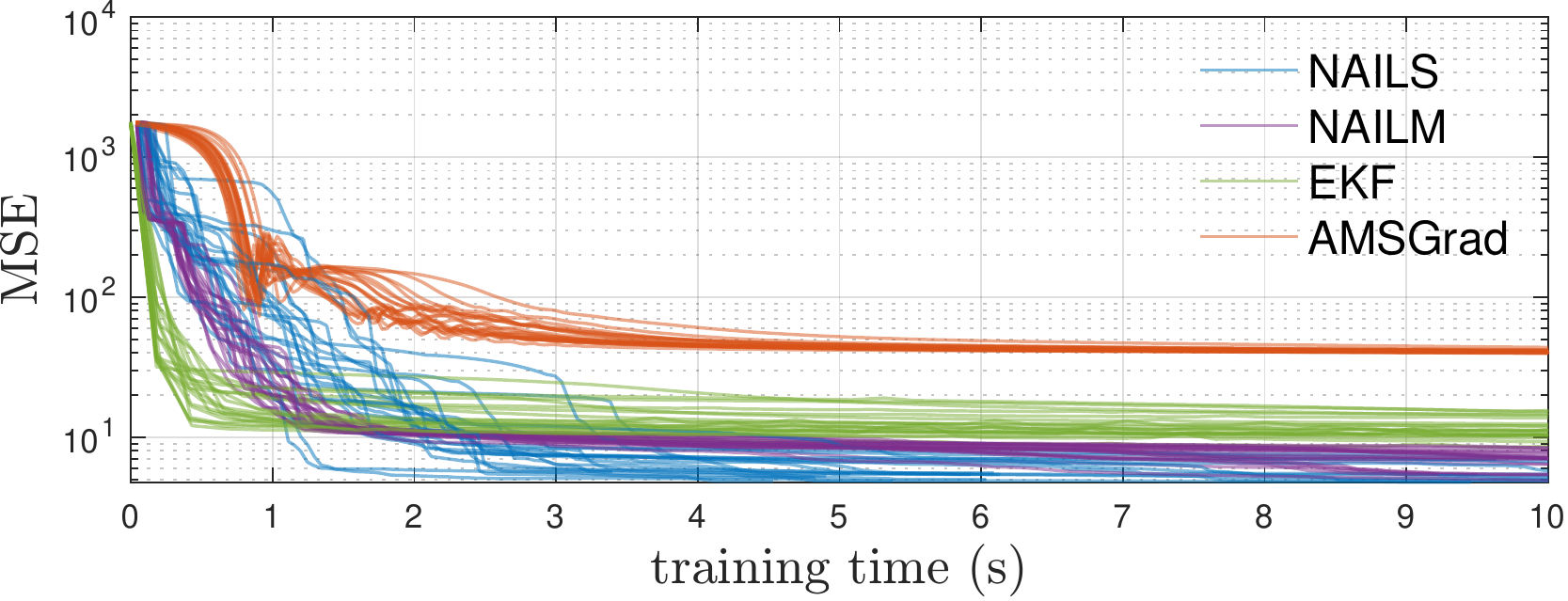}}
\caption{MSE loss~\eqref{eq:training-cost-condensed} evaluated on training data as a function of training time on different runs}
\label{fig:V_train_fluiddamper}
\end{figure}

\begin{table}[t]
    \begin{center}
        \begin{tabular}{l|l}\hline
        $L_x-1$ & number of hidden layers in $f_x$\\
        $(n_1^x,\ldots,n_{L_x-1}^x)$ & number of neurons per layer in $f_x$\\
        $n_{\tx}$ & dimension of parameter vector $\tx$\\\hline
        $L_y-1$ & number of hidden layers in $f_y$\\
        $(n_1^y,\ldots,n_{L_y-1}^y)$ & number of neurons per layer in $f_y$\\
        $f_{L_y}^y$ & nonlinear output function in $f_y$\\
        $n_{\ty}$ & dimension of parameter vector $\ty$\\\hline
        $\sigma_0$ & scaling factor used in random initialization~\cite{GB10}\\\hline
        $\rho_x$ & penalty on $\|x_0\|_2^2$ (or on $\|\txo\|_2^2$)\\
        $\rho_\theta$ & penalty on $\|\tx\|_2^2$, $\|\ty\|_2^2$\\
        $\tau_x$ ($\tau_y$)& penalty on $\|\tx\|_1$ ($\|\ty\|_1$) \\
        $\tau_g$ & group-Lasso penalty\\
        $\tau_x^0$ ($\tau_y^0$)& penalty on $\|\tx\|_0$ ($\|\ty\|_0$) \\\hline
        $E$ & max number of epochs when solving \\[-.2em]&the nonlinear LS problem\\
        $\epsilon_V$ & optimality threshold\\
        $c_1,\sigma,N_\sigma $ & line-search parameters\\
        $c_2,c_3,N_\lambda,\lambda_0$ & LM parameters\\\hline
        $N_{\rm ADMM}$ & number of ADMM iterations\\
        $\rho$ & ADMM parameter\\\hline
        \end{tabular}
    \end{center}
    \caption{Hyper-parameters of Algorithm~\ref{algo:NAILS} (NAILS/NAILM)}
    \label{tab:hyper-params}
\end{table}

\section{Numerical experiments}
\label{sec:experiments}
We test NAILS and NAILM on three nonlinear system identification
problems. The hyper-parameters of the 
Algorithm~\ref{algo:NAILS} are summarized in Table~\ref{tab:hyper-params}.
All computations have been carried out in 
MATLABR2022b on an Apple M1 Max CPU,
with the library CasADi~\cite{AGHRD19} used 
to generate Jacobian matrices via automatic differentiation. 
The scaling factor $\sigma_0=0.15$ is used to initialize the weights of the neural networks
in all the experiments. Unless the state encoder~\eqref{eq:fx0} is used, after training a model the best initial condition $x_0$
is computed by solving the small-scale non-convex simulation-error
minimization problem
\begin{equation}
    \min_{x_0}\rho_x\|x_0\|_2^2+\sum_{k=0}^{N_0}\ell(y_k,\hat y_k)
\label{eq:PSO_x0}
\end{equation}
by using the particle swarm optimizer (PSO) PSwarm~\cite{VV09}
with initial population of $2n_x$ samples, each component of $x_0$ 
constrained in $[-3,3]$, and with $N_0=100$. 

\subsection{RNN training with smooth quadratic loss}
\label{sec:example-1}
We first test NAILS/NAILM against the EKF approach~\cite{Bem21} and gradient-descent
based on the AMSGrad algorithm~\cite{RKK19}, for which we compute the gradient
as in~\eqref{eq:grad-V}, on real-world data from the nonlinear 
magneto-rheological fluid damper benchmark proposed in~\cite{WSCH09}. 
We use $N=2000$ data samples for training and $1499$ samples for testing, obtained from the System Identification (SYS-ID) Toolbox for MATLAB R2021a~\cite{Lju01}, where a nonlinear autoregressive (NLARX) model identification algorithm is employed for black-box nonlinear modeling. 

We consider a RNN model~\eqref{eq:RNN-xy} with $n_x=4$ hidden states and shallow state-update and output network functions ($L_x=L_y=2$) with $n_1^x=n_1^y=4$ neurons, hyperbolic-tangent  activation functions $f_1^x,f_1^y$, and linear output function $f_2^y$, parameterized
by $\tx\in\rr^{44}$, $\ty\in\rr^{29}$.
The CPU time to solve~\eqref{eq:PSO_x0} is approximately 30~ms in all tests.

In~\eqref{eq:training} we use the quadratic loss 
$\ell(y_k,\hat y_k)=\frac{1}{N}\|y_k-\hat y_k\|_2^2$ and quadratic regularization
$r(x_0,\tx,\ty)=\frac{1}{2}(\rho_x\|x_0\|_2^2+\rho_\theta\|\tx\|_2^2+\rho_\theta\|\tx\|_2^2)$
with $\rho_x=1$, $\rho_\theta=0.1$.
As $g(\tx,\ty)=0$, NAILS and NAILM only execute Algorithm~\ref{algo:SLS} once
with parameters $\epsilon_V=10^{-6}$ and, respectively, $c_1=10^{-4}$, $\sigma=0.5$, $N_\sigma=20$, and $\lambda_0=100$, $c_2=1.5$, $c_3=5$, $N_\lambda=20$. 
The initial condition is $x_0^0=0$ and the components of $\tx^0$, $\ty^0$ are randomly generated as described in Section~\ref{sec:initialization}.

Table~\ref{tab:fluid-damper-loss} shows the mean and standard deviation
(computed over 20 experiments, each one starting from a different random set of weights and zero bias terms) of the final best fit rate BFR = $100(1-\|Y-\hat Y\|_2/\|Y - \bar y\|_2)$, where $Y$ is the vector of measured output samples, $\hat Y$ the vector of output samples simulated by the identified model fed in open-loop with the input data, and $\bar y$ is the mean of $Y$,
achieved on training and test data.
\begin{table}[t]
    \begin{center}
        {\normalsize{
        \begin{tabular}{l|c|c}
        BFR & training & test\\[0em]\hline
        NAILS & 94.41 (0.27) & 89.35 (2.63)\\[0em]
        NAILM & 94.07 (0.38) & 89.64 (2.30)\\[0em]
        EKF & 91.41 (0.70) & 87.17 (3.06)\\[0em]
        SGD & 84.69 (0.15) & 80.56 (0.18)\\\hline
        \end{tabular}
        }}
    \end{center}
    \caption{Fluid damper benchmark: mean (standard deviation) of BFR
    on training and test data}
    \label{tab:fluid-damper-loss}
\end{table}
For illustration, the evolution of the MSE loss $\frac{1}{2N}\sum_{k=0}^{N-1}(y_k-\hat y_k)^2$
as a function of training time is shown in Figure~\ref{fig:V_train_fluiddamper}
for each run. The average CPU time per epoch spent by the training algorithm is 54.3~ms (NAILS), 72.2~ms (NAILM), 253.3~ms (EKF), and 34.6~ms (AMSGrad).

It is apparent that NAILS and NAILM obtain similar quality of fit and are computationally comparable. They get better fit results than EKF on average 
and, not surprisingly, outperform gradient descent due to their second-order approximation nature. On the other hand, EKF converges more quickly to a good quality of fit, as the model parameters $\tx$, $\ty$ are updated \emph{within} each epoch, rather than
\emph{after} each epoch is processed.

\subsubsection{Silverbox benchmark problem}
\label{sec:silverbox}
We test the proposed training algorithm on the Silverbox dataset, a popular benchmark for nonlinear system identification, for which we refer the reader to~\cite{WS13} for a detailed description. The first 40000 samples are used as the test set, the remaining samples, which correspond to a sequence of 10 different random odd multi-sine excitations, are manually split in $M=10$ different training traces of about 8600 samples each (the intervals between each different excitation are removed from the training set, as they bring no information).

We consider a RNN model~\eqref{eq:RNN-xy} with $n_x=8$ hidden states, no I/O feedthrough
($A_1^y\in\rr^{n_1^y\times n_x}$ in~\eqref{eq:RNN-y}), and state-update and output network functions with three hidden layers ($L_x=L_y=4$) with $8$ neurons
each, a neural network model $f_{x0}$~\eqref{eq:fx0} with two hidden layers of $4$
neurons each mapping the vector $v_0$ of past $n_a=8$ outputs and $n_b=8$ inputs
to the initial state $x_0$, hyperbolic-tangent activation functions, and linear output functions.
The total number of parameters is $n_{\tx}+n_{\ty}+n_{\txo}=296+225+128=649$, that we train using NAILM\footnote{The resulting model can be retrieved
at \url{http://cse.lab.imtlucca.it/~bemporad/shared/silverbox/rnn888.zip}} on $E=150$ epochs with $\rho_\theta=0.01$ and regularization $\frac{0.1}{2}\|\txo\|_2^2$
on the parameters defining $f_{x0}$.
For comparison, we consider the autoregressive models considered in~\cite{LZLJ04}, in particular the ARX model \textsf{ml}
and NLARX models \textsf{ms}, \textsf{mlc}, \textsf{ms}, and \textsf{ms8c50},
that we trained on the same dataset by using the SYS-ID Toolbox~\cite{Lju01}
using the same commands reported in~\cite{LZLJ04}. Note that the latter two models contain explicitly the nonlinear term $y_{k-1}^3$, in accordance with the physics-based model of the Silverbox system, which is an electronic implementation of the Duffing oscillator. The results of the comparison are shown in Table~\ref{tab:silverbox}, which also includes the results reported 
in the recent paper~\cite{BTS21} and those obtained by the LSTM model proposed in~\cite{LATS20}. The results reproduced here differ from those reported originally in the papers, that we show included in brackets (the results originally reported in~\cite{LATS20} have been omitted,
as they were obtained on a reduced subset of test and training data).
\begin{table}[t]
    \begin{center}
    {\normalsize{
    \begin{tabular}{l|c|c}
        identification method &RMSE (mV) & BFR 
        \\[0em] \hline
        ARX (ml)~\cite{LZLJ04} & 16.29 [4.40]& 69.22 [73.79]\\[0em] 
        NLARX (ms)~\cite{LZLJ04} & \phantom{0}8.42 [4.20]& 83.67 [92.06]\\[0em]
        NLARX (mlc)~\cite{LZLJ04} & \phantom{0}1.75 [1.70]& 96.67 [96.79]\\[0em]        
        NLARX (ms8c50)~\cite{LZLJ04} & \phantom{0}1.05 [0.30] & 98.01 [99.43]\\[0em]
        Recurrent LSTM model~\cite{LATS20} & \phantom{0}2.20 \phantom{[5.9\phantom{0}]}& 95.83 \phantom{[88.9\phantom{0}]}\\[0em]
        State-space encoder~\cite{BTS21}~($n_x=4$) &\phantom{00.00} [1.40] & \phantom{00.00} [97.35]\\[0em]
        NAILM (this paper) & 0.35 \phantom{[0.00]}& 99.33 \phantom{[00.00]}
        \\[0em] \hline
    \end{tabular}
    }}\end{center}
    \caption{Silverbox benchmark: RMSE and BFR obtained by different methods. The numbers reported in brackets refer to the results reported in the corresponding original papers.}
    \label{tab:silverbox}
\end{table}

\subsection{RNN training with $\ell_1$-regularization}
\label{sec:example-2}
\begin{figure}[t]
\centerline{\includegraphics[width=0.8\hsize]{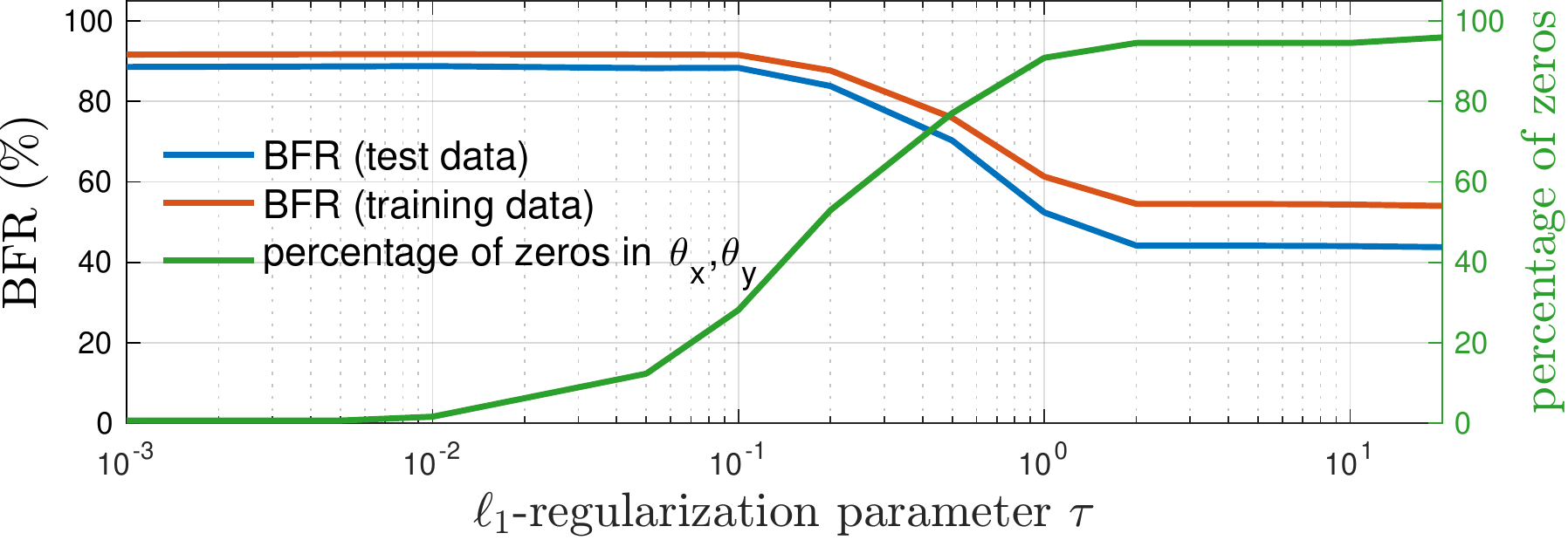}}
\caption{Fluid damper benchmark: BRF and sparsity of $\tx,\ty$
for different $\ell_1$-regularization coefficients $\tau$
(mean results over 20 runs from different random
initial weights)}
\label{fig:L1_fluiddamper}
\end{figure}

We consider again the fluid damper benchmark dataset and add 
an $\ell_1$-regularization term $g(\tx,\ty)$ as in~\eqref{eq:L1-reg-fun} and use NAILS to train the same RNN model structure~\eqref{eq:RNN-xy},
for different values of $\tau_x=\tau_y\eqdef\tau$. Figure~\ref{fig:L1_fluiddamper} shows the resulting BRF.
Figure~\ref{fig:L1_fluiddamper} also shows the percentage of zero coefficients in the resulting parameter vector $\smallmat{\tx\\\ty}\in\rr^{73}$, i.e., its sparsity.

The benefit of using $\ell_1$-regularization to reduce the complexity of the model
without sacrificing fit quality is clear. In fact, the BRF value
on test data remains roughly constant until $\tau\approx 0.1$, which corresponds to 
$\approx 30\%$ zero coefficients in the model.

In order to compare NAILS/NAILM with other state-of-the-art training algorithms
that can deal with $\ell_1$-regularization terms
(Adam~\cite{KB14}, AMSGrad~\cite{RKK19}, DiffGrad~\cite{DCRMSC19}), 
we report in Table~\ref{tab:l1_sgd_vs_nails}
the results obtained by solving the training problem for $\tau=0.2$
using the EKF approach~\cite{Bem21} and some of the SGD
algorithms mostly used in machine-learning packages. NAILS/NAILM is run for $E=250$, 
SGD for $E=2000$, and EKF for $E=50$ epochs. While all methods get a similar BFR, 
NAILS and NAILM are more effective in both sparsifying the model-parameter vector
and execution time.

\begin{table}[t]
\begin{center}{\normalsize{
\begin{tabular}{l|c|c|c|c}
training & BFR & BFR & sparsity & CPU \\[0em]
algorithm & training & test & \% & time (s)\\[0em]\hline
NAILS & 91.00 (1.66) & 87.71 (2.67) &  65.1 (6.5) &  11.4\\[0em]
NAILM & 91.32 (1.19) & 87.80 (1.86) &  64.1 (7.4) &  11.7\\[0em]
EKF & 89.27 (1.48) & 86.67 (2.71) &  47.9 (9.1) &  13.2\\[0em]
AMSGrad & 91.04 (0.47) & 88.32 (0.80) &  16.8 (7.1) &  64.0\\[0em]
Adam & 90.47 (0.34) & 87.79 (0.44) &   \phantom{0}8.3 (3.5) &  63.9\\[0em]
DiffGrad & 90.05 (0.64) & 87.34 (1.14) &   \phantom{0}7.4 (4.5) &  63.9\\[0em]
\hline
\end{tabular}
}}\end{center}
\caption{Fluid damper benchmark: mean BRF, sparsity of $[\tx'\ \ty']'$
(with standard deviation in parentheses), and mean CPU time given by different solution methods
when solving $\ell_1$-regularized training for $\tau=0.2$
(results obtained over 20 runs from different random initial weights)}
\label{tab:l1_sgd_vs_nails}
\end{table}

\begin{figure}[t]
\centerline{\includegraphics[width=0.8\hsize]{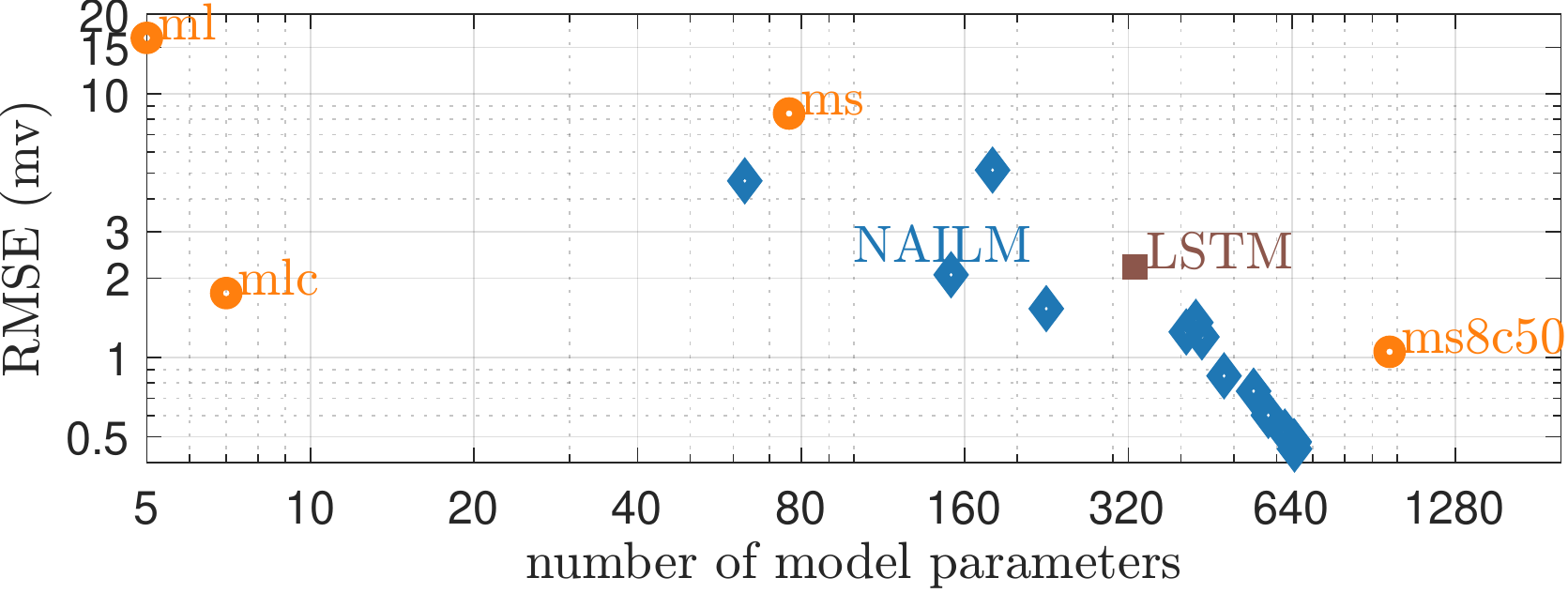}}
\caption{Silverbox benchmark: RMSE vs number of parameters. Models \textsf{ml},
\textsf{ms}, \textsf{mlc}, and \textsf{ms8c50}~\cite{LZLJ04},
LSTM model~\cite{LATS20}, and RNN models generated by NAILM for values of
the $\ell_1$-regularization parameter $\tau$ between 0 and 20.}
\label{fig:num_params}
\end{figure}

For the Silverbox benchmark problem, following the analysis performed in~\cite{MSRS13}, 
Figure~\ref{fig:num_params} shows the obtained RMSE on test data as a function of the number of nonzero parameters in the model that are obtained by running NAILM to train the same RNN model structure defined in Section~\ref{sec:silverbox} under an additional $\ell_1$-regularization term with weight $\tau$ between $10^{-4}$ and 20.
The figure shows the effectiveness of NAILM in trading off model simplicity vs quality of fit.

\subsection{RNN training with quantization}
\label{sec:example-3}
Consider again the fluid damper benchmark dataset and let $\tx$, $\ty$ be only allowed to take values in the finite set $\QQ$ of the multiples of $0.1$ 
between $-0.5$ and $0.5$.
We also change the activation function of the neurons
to the leaky-ReLU function $f(x)=\max\{x,0\}+0.1\min\{x,0\}$,
so that the evaluation of the resulting model amounts to extremely simple
arithmetic operations, and increase the number of hidden neurons to $n_1^x=n_1^y=6$. 
We train the model using NAILS to handle the non-smooth
and nonconvex regularization term $g(\tx,\ty)=0$ if $\tx\in\QQ^{n_{\tx}}$,
$\ty\in\QQ^{n_{\ty}}$, $g(\tx,\ty)=+\infty$ otherwise, with $E=1$,
$N_{\rm ADMM}=200$, $\rho=10$, without changing the remaining parameters.
For comparison, we solve the same problem without non-smooth regularization
term $g$ by running Algorithm~\ref{algo:SLS} for $E=N_{\rm ADMM}$ epochs and then
quantize the components of the resulting vectors $\tx$, $\ty$ to their closest values in $\QQ$.

The mean and standard deviation of the BRF and corresponding CPU time
obtained over 20 runs from different random
initial weights is reported in Table~\ref{tab:quantization}.
It is apparent that quantizing a posteriori the parameters of a trained RNN leads to
much poorer results.
\begin{table}[t]
    \begin{center}
    {\normalsize{
        \begin{tabular}{l|c|c}
        BFR & NAILS & quantization\\[0em]\hline
        training & 84.36 (3.00) &17.64 (6.41)
        \\[0em]test & 78.43 (4.38) &12.79 (7.38)\\[0em]\hline
        CPU time & 12.04 (0.54) & \phantom{0}7.19 (2.82)
        \\[0em]\hline
        \end{tabular}
    }}
    \end{center}
    \caption{Fluid damper benchmark with quantized model
    coefficients: mean BRF and CPU time (standard deviation)
    over 20 runs}
    \label{tab:quantization}
\end{table}

\subsection{Group-Lasso penalty and model-order reduction}
\label{sec:example-4}
\begin{figure}[t]
\centerline{\includegraphics[width=0.8\hsize]{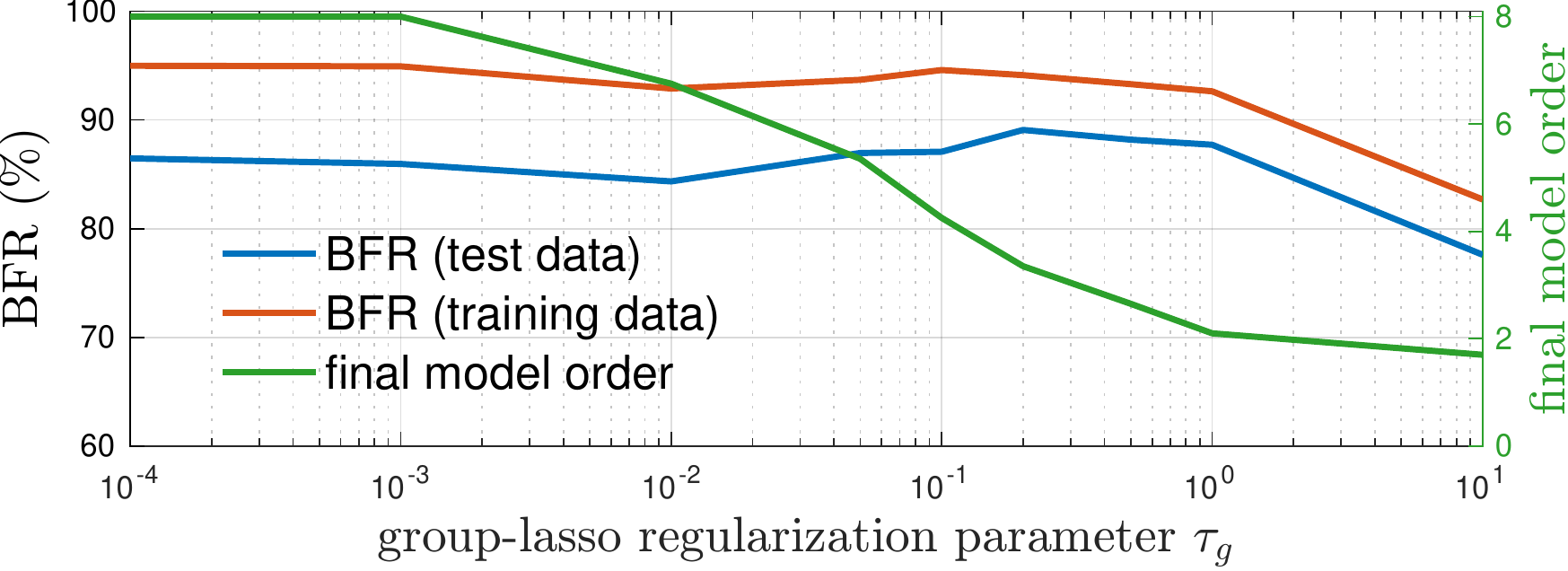}}
\caption{BRF and resulting model order $n_x$
for different group-Lasso regularization coefficients $\tau_g$
(mean results over 20 runs from different random
initial weights)}
\label{fig:L1_fluiddamper-groupLasso}
\end{figure}

We use the group-Lasso penalty $\tau_g$ described in
Section~\ref{sec:groupLasso} to control the model order $n_x$
when training a model on the fluid damper benchmark dataset.
We consider the same settings as in Section~\ref{sec:example-1},
except for the model-order $n_x=8$, $n_1^x=n_1^y=6$  neurons in
the hidden layers of the state-update and output functions, and set
the ADMM parameter $\rho=10\tau_g$. 
The mean BRF results over 20 different runs of the NAILS algorithm
obtained for different values of $\tau_g$ are reported in Figure~\ref{fig:L1_fluiddamper-groupLasso}, along with the resulting mean model order obtained. From the figure, one can see that $n_x=3$ is a good candidate model order to provide the best
BRF on test data.

\subsection{RNN training with smooth non-quadratic loss}
To test the effectiveness of the proposed approach in handling non-quadratic
strongly convex and smooth loss terms, we consider $2000$ input/output pairs generated by the following nonlinear system with binary outputs
\[
    \ba{rcl}
    x(k+1)&=&\smallmat{.8 & .2 & -.1\\ 0 & .9 & .1\\ .1 & -.1 & .7}\diag(x(k))(0.9+
    0.1\sin(x(k)))\\[.5em]&&+
             \smallmat{-1\\.5\\1}u(k)(1-u^3(k))+\xi(k)\\[.5em]
    y(k)&=&\left\{\ba{ll}1&\mbox{if}\ {\scriptsize\matrice{c}{-2 \\[-.3em] 1.5 \\[-.3em] 0.5}}'
    \hspace*{-.5em}(x(k)+\frac{1}{3}x^3(k))-{\scriptsize{4}}+\zeta(k)\geq 0\\0&\mbox{otherwise}\ea\right.
    \ea
\]
from $x(0)=0$, with the values of the input $u(k)$ changed with 90\% probability
from step $k$ to $k+1$ with a new value drawn from the uniform distribution on $[0,1]$. 
We consider independent noise signals $\xi_i(k),\zeta(k)\sim \NN(0,\sigma_n^2)$, $i=1,2,3$, for
different values of the standard deviation $\sigma_n$. The first $N=1000$ samples are used
for training, the rest for testing. NAILS and NAILM are used to train a RNN model~\eqref{eq:RNN-xy} with no feedthrough, one hidden layer ($L_x=L_y=2$) with $n_1^x=n_1^y=5$ neurons and hyperbolic-tangent activation function, and sigmoid output function $f_2^y(y)=1/(1+e^{-A_2^y[x'(k)\ u(k)]'-b_2^y})$. The resulting model-parameter vectors $\tx\in\rr^{43}$ and $\ty\in\rr^{26}$ are trained using the
modified cross-entropy loss $\ell(y(k),\hat y)=\sum_{i=1}^{n_y} -y_i(k)\log(\epsilon+\hat y_i)-(1-y_i(k))\log(1+\epsilon-\hat y_i)$~\cite{Bem21}, where $\hat y=f_y(x_k,\ty)$ and $\epsilon=10^{-4}$, quadratic regularization with $\rho_x=0.1$, $\rho_\theta=0.01$,
optimality tolerance $\epsilon_V=10^{-6}$, parameters $c_1=10^{-4}$, $\sigma=0.5$, $N_\sigma=10$ (NAILS), and $\lambda_0=100$, $c_2=1.5$, $c_3=5$, $N_\lambda=30$ (NAILM),
for maximum $N_{\rm ADMM}=150$ epochs.

Table~\ref{tab:linear-binary} shows the final accuracy (\%) achieved on training
and test data for different levels $\sigma_n$ of noise, averaged over 20 runs from different initial values of the model parameters. The average CPU time to converge is 5.2~s (NAILS)
and 6.2~s (NAILM).

\begin{table}[t]
    \begin{center}
        {\normalsize{
\begin{tabular}{cl|c|c}
$\sigma_n$ && accuracy (\%) & accuracy (\%) \\[0em]
 & & training data & test data\\[0em]\hline
0.00 & NAILS & 99.16 (0.6) & 96.87 (1.0)\\[0em]
& NAILM & 99.30 (1.0) & 96.66 (1.3)\\\hline
0.05 & NAILS & 98.45 (0.7) & 94.53 (1.2)\\[0em]
& NAILM & 98.13 (0.3) & 94.73 (1.9)\\\hline
0.20 & NAILS & 86.03 (4.1) & 83.32 (5.5)\\[0em]
& NAILM & 86.33 (1.5) & 85.52 (1.7)\\\hline
\end{tabular}
        }}
\end{center}
\caption{Nonlinear system with binary outputs: mean final accuracy (\%) (standard deviation)
over 20 runs on training and test data}
\label{tab:linear-binary}
\end{table}
\section{Conclusions}\label{sec:conclusions}
We have proposed a training algorithm for RNNs that is computationally efficient,
provides very good quality solutions, and can handle rather general loss and regularization terms. Although we focused on recurrent neural networks of the form~\eqref{eq:RNN-xy}, 
the algorithm is applicable to learn parametric models
with rather arbitrary structures, including other black-box state-space models (such as LSTM models), gray-box and white-box models, and static models.

Future research will be devoted to further analyze the convergence properties of
Algorithms~\ref{algo:SLS} and~\ref{algo:LM} within the Generalized Gauss-Newton framework, and
to establish conditions on the ADMM parameter $\rho$, number of epochs $E$ used
when solving~\eqref{eq:ADMM-x}, the model structure to learn, and the regularization function $g$ that guarantee convergence of the nonconvex ADMM iterations.

\section{Acknowledgement}
The author thanks M. Schoukens for exchanging ideas on the setup of the Silverbox benchmark problem.

\newcommand{\noopsort}[1]{}

\end{document}